\documentclass{article}

\usepackage[final]{corl_2020} 
\usepackage{float}
\usepackage{cite}
\usepackage{graphics,graphicx}
\usepackage{xcolor}
\usepackage{soul}
\usepackage{amsmath,mathtools}
\usepackage[utf8]{inputenc}
\usepackage[acronym,smallcaps]{glossaries}
\usepackage{subfigure}
\usepackage{booktabs}

\newcommand{\fref}[1]{Fig. \ref{#1}}
\newcommand{\tref}[1]{Table \ref{#1}}

\newcommand{\sref}[1]{Section \ref{#1}}

\newacronym{rl}{RL}{\textit{Reinforcement Learning}}
\newacronym{mfrl}{MFRL}{\textit{Model-Free Reinforcement Learning}}
\newacronym{ml}{ML}{\textit{Machine Learning}}
\newacronym{marl}{MARL}{\textit{Multi-agent Reinforcement Learning}} 
\newacronym{il}{IL}{\textit{Independent Learners}} 
\newacronym{jal}{JAL}{\textit{Joint-Action Learners}} 

\newacronym{vsss}{VSSS}{\textit{Very Small Size Soccer}}
\newacronym{ssl}{SSL}{\textit{Small Size League}}
\newacronym{mdp}{MDP}{\textit{Markov Decision Process}}
\newacronym{larc}{LARC}{\textit{Latin American Robotics Competition}}

\newacronym{dqn}{DQN}{\textit{Deep Q Network}}
\newacronym{ddpg}{DDPG}{\textit{Deep Deterministic Policy Gradient}}
\newacronym{sac}{SAC}{\textit{Soft Actor Critic}}
\newacronym{drl}{DRL}{\textit{Deep Reinforcement Learning}}
\newacronym{fnn}{FNN}{\textit{Fuzzy Neural Network}}
\newacronym{ac}{AC}{\textit{Actor Critic}}

\newacronym{model}{VSSS-RL}{\textit{Very Small Size Soccer Reinforcement Learning}}

\title{
A Framework for Studying Reinforcement Learning and Sim-to-Real in Robot Soccer
}

\author{Hansenclever F. Bassani$^{1,2}$, Renie A. Delgado$^{1}$, José Nilton de O. Lima Junior$^{1}$,\\\textbf{Heitor R. Medeiros$^{1}$, Pedro H. M. Braga$^{1}$, Mateus G. Machado$^{1}$,}\\\textbf{Lucas H. C. Santos$^{1}$, Alain Tapp$^{2}$}
\thanks{$^{1}$Centro de Informática - Universidade Federal de Pernambuco, Av. Jornalista Anibal Fernandes, s/n - CDU 50.740-560, Recife, PE, Brazil. Corresponding author: Hansenclever Bassani {\tt\small hfb@cin.ufpe.br}}
\thanks{$^{2}$Mila, Universite de Montréal, Montréal, Québec, Canada, H3C 3J7}%
}
\begin{document}

\maketitle

\thispagestyle{empty}
\pagestyle{empty}

\begin{abstract}
This article introduces an open framework, called VSSS-RL, for studying \acrfull{rl} and sim-to-real in robot soccer, focusing on the IEEE \acrfull{vsss} league. We propose a simulated environment in which continuous or discrete control policies can be trained to control the complete behavior of soccer agents and a sim-to-real method based on domain adaptation to adapt the obtained policies to real robots. Our results show that the trained policies learned a broad repertoire of behaviors that are difficult to implement with handcrafted control policies. With VSSS-RL, we were able to beat human-designed policies in the 2019 \acrfull{larc}, achieving 4th place out of 21 teams, being the first to apply \gls{rl} successfully in this competition. Both environment and hardware specifications are available open-source to allow reproducibility of our results and further studies.
\end{abstract}



\section{Introduction}


Every year, the \acrfull{larc} promotes the IEEE \gls{vsss}, a traditional robot soccer competition in which two teams of three small differential drive robots compete to score goals against each other (\fref{fig:vss_sim}). In the \gls{vsss} league, the robots are typically programmed to behave adequately in every situation identified by the programmers employing path planning, collision avoidance, and PID control methods\citep{kim2004soccer}. However, it is tough to foreseen and tackle every possible situation in an unpredictable game such as soccer, limiting what can be achieved by hard-coded behaviors.

\gls{rl} gained popularity when it became capable of handling increasingly complex decision-making problems in simulated environments, such as learning how to play Atari games \citep{volodymyr2013playing}, Chess \citep{campbell2002deep}, and Starcraft 2 \citep{vinyals2019alphastar}, achieving human-level performance. 

In robotics, \gls{rl} showed promising results in simulated and real-world environments, including approaches for motion planning, optimization, grasping, manipulation, and control \citep{christiano2016transfer,andrychowicz2018learning,levine2016end}. More specifically, in the literature of robot soccer, \gls{rl} has been applied for learning specific behaviors, such as kicking and scoring goals \citep{duan2007application,zhu2019sslrl}. However, obtaining control policies for the complete behavior of robots playing soccer in the real world is still an open problem, even in the simplest categories, such as \gls{vsss}. In the real world, several barriers exist to obtain good results with \gls{rl}, as discussed in \citep{dulac2019challenges}. For instance, the large amounts of interactions required by the agents to achieve adequate performance are frequently impractical due to the degradation of hardware, energy consumption, and time.

An alternate approach for this problem is training in simulation and transfer the learned policy to the real world, which is known as sim-to-real. As simulations are, by definition, an approximation of the real world, there is a reality gap between simulated and real environments, i.e., intrinsic discrepancies, such as friction, gear backslash, sensor and actuators noise, delays, misaligned and deformable robot parts. \gls{rl} methods are known to be optimistically biased, i.e., tend to overfit to the simulate environment. Therefore, these models will perform poorly in the real world as the reality gap degrades its performance. Sim-to-real methods seek to minimize the reality gap by either trying to approximate the simulated environment to the real configuration or trying to produce more generalized policies.

Two \gls{rl} open soccer environments have been proposed: MuJoCo Soccer \citep{todorov2012mujoco} and Google Research Football \citep{kurach2019google}. However, they are not suitable for the study of sim-to-real, as they either do not consider important physical and dynamical aspects or represent a very complex scenario that is not achievable by current robotics technology.

Considering this, the contributions of the present work are: (i) an open framework that can be used as a benchmark tool for the community studying \gls{rl}, multi-agent \gls{rl}, and sim-to-real in dynamic, competitive and cooperative scenarios. The framework includes the hardware specifications of our inexpensive robots and an OpenAI Gym \citep{gym} environment which can interface with both a \gls{vsss} simulator and the real-world robots; (ii) the evaluation of the performance of three baseline \gls{rl} methods for training the agents in simulation with discreet and continuous actions; (iii) a sim-to-real approach based on a feed-forward neural network to create an abstraction layer between high-level and low-level control commands. Both environment and hardware specifications are available open-source, aiming at making our results easily reproducible by others, avoiding the reproducibility issues that we have observed in the field. 

The results show that, in the single-agent scenario, this approach can achieve a fine level of control and learn the complete behavior of a general-purpose \gls{vsss} league agent. The obtained policy was able to match the level of the polices designed and refined by humans. Thus, by replicating the best single agent policy for three agents to compete in the 3-vs-3 game setup, our team achieved fourth place out of 21 teams in \gls{larc} 2019, being the first to successfully apply \gls{rl} in this competition.

The rest of this article is organized as follows: \sref{sec:related} presents related work on \gls{rl} for robot soccer. \sref{sec:environment} presents the proposed framework, discussing the adaptations required in the simulator and the Gym wrappers created. \sref{sec:approach} describes the method we used to transfer the policies learned in simulation to the real world. \sref{sec:results} presents the results, and finally, \sref{sec:conclusion} draws the conclusions and proposes future work.

\section{Related Work}
\label{sec:related}

Commonly, \gls{rl} is used in robot soccer to learn a set of desired skills. In \citep{riedmiller2009reinforcement}, Batch Reinforcement Learning methods are applied for robot soccer to steal the ball of a player and to perform low-level motor control. The method samples experiences in the real world, generates training patterns dynamically, and approximates the function represented by them through batch supervised learning.


Moreover, in \citep{yoon2017new}, an \gls{rl} approach is proposed for learning certain skills for RoboCup \gls{ssl} soccer robots. In particular, it focuses on infinite \gls{mdp} problems, in which the dynamics of the environment is known. The approach is applied for learning shooting skills under a variety of different scenarios.

In \citep{catacora2019cooperative}, two different \gls{marl} approaches are used for a 2-vs-2 free-kick task on a physically realistic 3D simulator: \gls{il}, and \gls{jal}. In the first, every agent performs standard \gls{rl}, but in the presence of other agents, whereas in the latter, the state and action spaces of all agents are merged. So, just a single policy is learned to map joint-observations to joint-actions.

It is also important to distinguish the concept of learning at a high or at a low level of abstraction. For instance, the authors of \citep{abreuhigh2019learning} and \citep{abreulow2019learning} separate this in two different applications in a simulated environment. First, in \citep{abreuhigh2019learning}, the action space comprises two controllable, abstract, and consequently discrete commands: \textit{dash}, to get closer to the ball, and \textit{kick}, to push the ball. Second, in \citep{abreulow2019learning}, the objective is to produce continuous actions, which are considered as low-level.

Most works on robot soccer aim at learning specific skills, instead of the complete behavior of a soccer agent. Examples can be pinpointed, as in \citep{riedmiller2007experiences} for kicking or in \citep{hester2010generalized} for scoring penalty goals. In \citep{duan2007application}, the authors present a hierarchical \gls{rl} approach in a context similar to this paper. They combined Q-learning with \gls{fnn} to provide a learning approach for decision making in the robot soccer domain. The authors choose to divide the task hierarchically in multiple independent sub-tasks. The sub-tasks learned are divided as follows: learn to shoot, run off the ball, role assignment, and action selection between the learned skills and a set of hand-designed behaviors. The \gls{fnn} maps the state space into a continuous action space for Q-learning. The approach was capable of learning in both the simulation and the real world. The solution was victorious in the 7th Robot Soccer Tournament 5-vs-5 runner-up and in the China FIRA championship. Although the paper reaches remarkable results, it requires a great amount of human engineering. 

While in \citep{duan2007application} the authors focus on a hierarchical approach in a similar context to ours, we choose to investigate an end-to-end approach in robot soccer. The objective is to develop skills that emerge directly from the training of the agent with state-of-art \gls{mfrl} algorithms in the simulated environment. Also, we seek behaviors that suitably transfer to the real world.

Regarding sim-to-real, in Domain Randomization, a policy is trained on a set of environments with randomized parameters to improve the robustness of the agent to these variable environmental factors \citep{andrychowicz2018learning}. One of the main drawbacks of this approach is the high amounts of samples needed for the agent to generalize to the environment variations. Some works try to decrease the number of samples needed by minimizing the set of environment variations, adjusting the random distribution using real-world roll-outs \citep{chebotar2018closing}. In Domain Adaptation, the actions taken by the policy (learned in simulation) are mapped to the corresponding actions in the real environment, aiming to produce a similar outcome.

Certain works try to improve the environment distribution by using real-world data to provide better quality samples \citep{cubuk2018autoaugment}. Other approaches can be indirectly linked to sim-to-real. For instance, \gls{rl} algorithms can be developed to take into consideration robotics limitations, such as safety constraints \citep{haarnoja2018soft}, or to infer world models from data, aiming to produce canonical representations \citep{golemo2018sim, higgins2017darla}.



\section{VSSS-RL Framework}
\label{sec:environment}

The proposed framework, VSSS-RL, is an \gls{rl} framework that allows the study and application of diverse aspects of \gls{rl} in robot soccer, such as cooperation, reward assignment, competition, and sim-to-real. It uses a modified version of the FIRA Simulator (FIRASim) \citep{firasim} or VSS-SDK \citep{vss_sdk} and builds a set of wrapper modules to achieve compatibility with the OpenAI Gym standards \citep{brockman2016openai}\footnote{Source code available at \href{url}{https://github.com/robocin/vss-environment}}.

\begin{figure*}[hb!]
  \centering
  \subfigure[]{\includegraphics[width=0.7\linewidth]{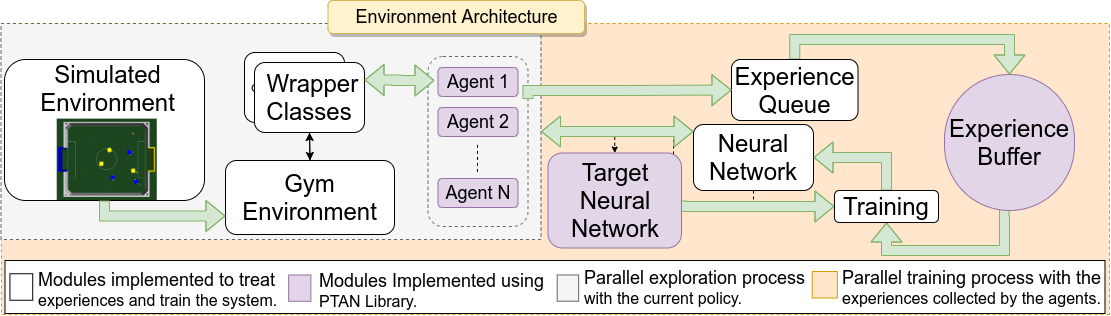}
  \label{fig:architecture}}
  \subfigure[]{\includegraphics[width=0.28\linewidth]{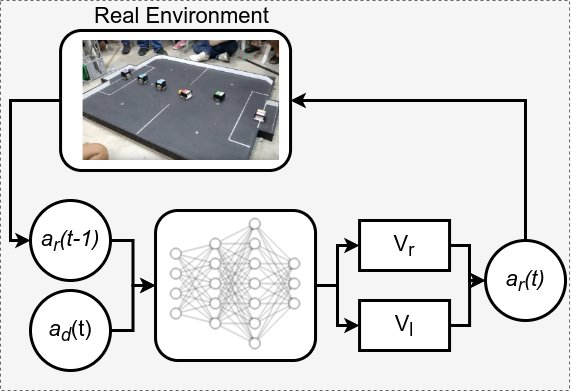}
  \label{fig:sim2realtrain}}

  \caption{VSSS-RL: (a) Architecture of \gls{vsss}-RL Environment: Experience and Training Processes for learning high-level control policies. (b) Low-level control training processes to enable the sim-to-real transfer. $a_{r}(t-1)$ are the linear and angular speeds observed in the previous step, $a_{d}(t)$ the action desired by the high-level policy, and $a_r(t)$ the action that should be taken in the real environment, in terms of right and left wheel speeds, $V_r$ and $V_l$.}
  \label{fig:vssss-env}
\end{figure*}

\gls{vsss}-RL consists of two main independent processes: 1) the experience process; and 2) the training process. In the first, an OpenAI Gym environment parser was developed, and wrapper classes were implemented to communicate with the agents. In the latter, the collected experiences are stored in an experience buffer, which is used to update the policies, as illustrated in \fref{fig:vssss-env}.

\subsection{Simulated}
\label{sec:simulator}
\label{sec:wrapper}

\begin{figure*}[ht!]
  \centering
  \subfigure[]{\includegraphics[width=0.215\linewidth,scale=1]{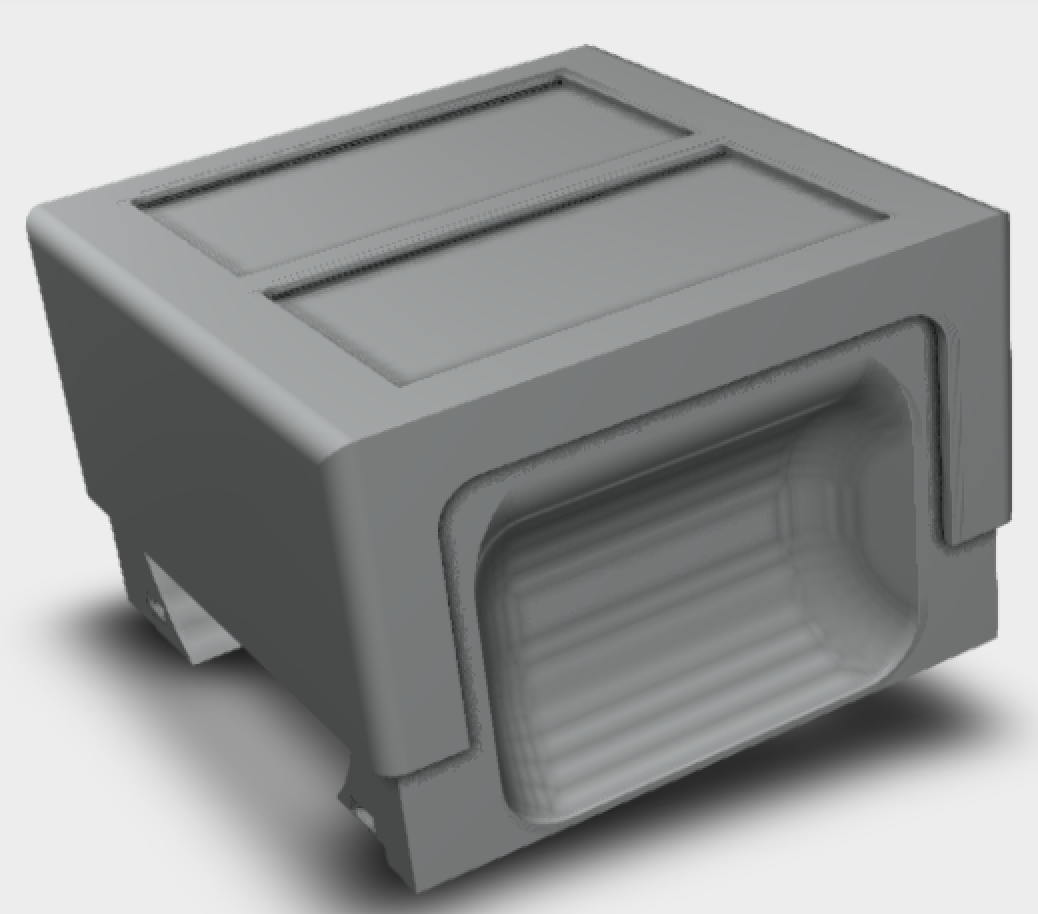}}
  \subfigure[]{\includegraphics[width=0.32\linewidth,scale=1]{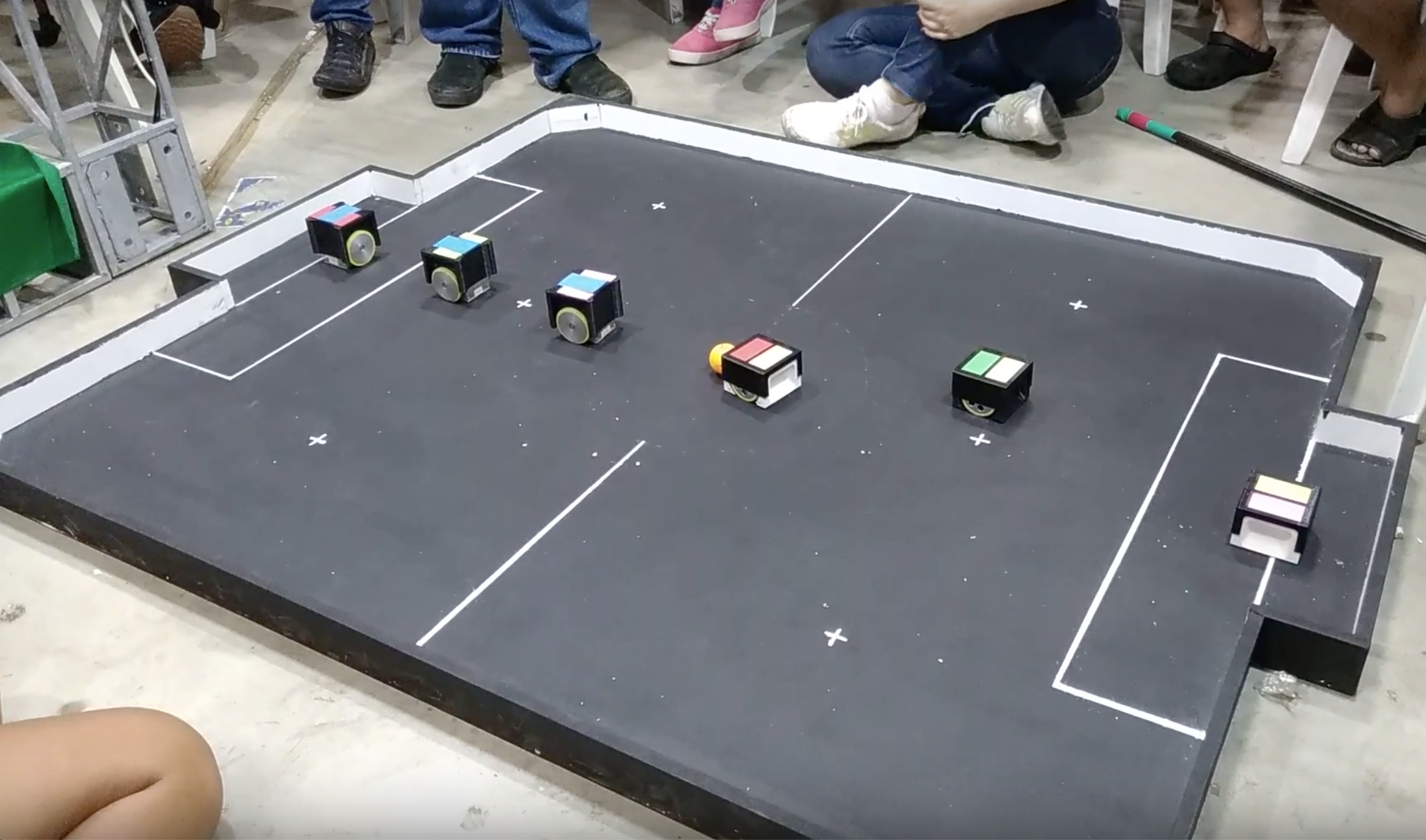}}
  \subfigure[]{\includegraphics[width=0.232\linewidth,scale=1]{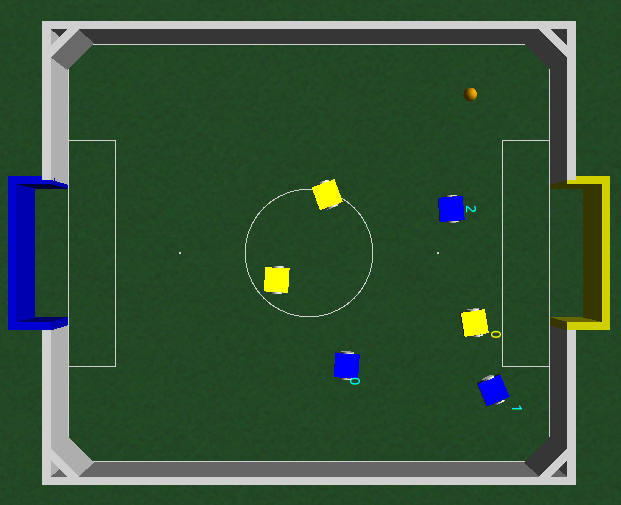}
    \label{fig:vss_sim_env}}
  \caption{3D model of a \gls{vsss} real robot(A) and the Real-world game setup (B). Visualization of FIRA simulation \citep{firasim} (3).}
  \label{fig:vss_sim}
\end{figure*}

The simulation is composed of a simulator and an environment. The FIRASim \citep{firasim} or the VSS-SDK \citep{vss_sdk}. Both are 3D simulators implemented in C++, using Open Dynamics Engine \citep{ode} to provide the \gls{vsss} environment and a view window (\fref{fig:vss_sim_env}) to  display the scenes. FIRASim is adapted to \gls{vsss} from grSim, a simulator of Robocup Small Size League.

The communication with the simulators is performed via sockets. The commands for each agent are composed of linear velocities of both wheels. The simulator state is then returned with the poses and velocities of the elements in the field. Two main modifications were made to adapt FIRASim and VSS-SDK for \gls{rl}: 1) The simulator was synchronized with the agents to decrease observation noise; and 2) Command-line parameters were introduced to setup the simulation (ports and frame rate).

Moreover, a Gym environment was developed to encapsulate the simulators. It is an API developed by OpenAI that aims to create a unified interface between the agents and different types of environments, e.g., discrete or continuous, real or simulated. In our environment, the observation is a vector of the pose and velocity of the ball and all the robots in the field plus a timestamp value in [0,1]. The actions of the agents consist of setting the linear speeds (values in the $[-100,100]$ interval) of both wheels, for each robot. An episode in our environment lasts 5 minutes of simulation time (half time of a real game). 



\subsection{Real World}

The setup for the \gls{vsss} real-world environment follows the standards presented in \citep{vss_rules}, which consists of a game field of $170$cm x $150$cm dimensions, and a camera (see \tref{tab:table_camera} for specifications) positioned $2$m above the field to capture robots and ball poses. The robots, illustrated in \fref{fig:vss_sim}, are designed accordingly to \tref{tab:table_robot}\footnote{3D Models are available at \href{url}{https://github.com/robocin/vss-mechanics/wiki}}. 
The chassis and wheels are 3D printed and the tires are made using silicon with 20 shores. The electronics uses widely available and inexpensive controllers and drivers. This combination of materials and techniques produces a simple robot with an adaptable design costing around USD \$130.00 per unity.


\begin{table}[H]
\parbox{.4\linewidth}{
    \centering
    \caption{Camera specifications}
    \begin{tabular}{|l|c|}
    \hline
        Model & Logitech C920 Pro \\
        \hline
        Resolution & $640\: x\: 480px$ \\
        \hline
        Frame rate &  $30 fps$\\
        \hline
        Interface & USB 2.0 \\
        \hline
        Latency & $90ms \pm 10ms $\\
        \hline
    \end{tabular}
    \label{tab:table_camera}
}
\parbox{.4\linewidth}{
    \caption{Robot specifications}
    \begin{tabular}{|l|l|}
        \hline
        Weight & 150g \\
        \hline
        Dimensions & $7,5cm \: x\: 7,5cm \: x\: 5,6cm$ \\
        \hline
        Wheel Radius &  $2,6 cm$\\
       \hline
        Microprocessor & $2x$ ATmega328 \\
        \hline
        Communication & Nordic nRFL2401+\\
        \hline
        Motors & $2x$ Micro Metal 50:1 6V\\
        \hline
        Motors Driver & TB6612FNG Dual Motor\\
        \hline
        Battery & $2x$ Lipo 300mA 2S\\
        \hline
    \end{tabular}
    \label{tab:table_robot}
}
\end{table}

An example of the game setup can be seen in \fref{fig:vss_sim}. The \gls{ssl} Vision software \citep{zickler2009ssl} is used to extract information about robots and ball poses from the frames captured by the camera. The vision software uses a color segmentation pipeline and translates the positions in the image to positions in the field plane. This data is then sent through the UDP packet and is received by our VSSS-RL environment in another process. To control the robots, the VSSS-RL sends wheel speeds via serial communication to its embedded board using a radio. The communication between computer and robots is performed using a broadcast network made of nRF24l01+ radios of 2.4 GHz, with a delay of $300\mu$s and a loss package rate of $0.08\%$. 
\section{High-Level and Low-Level Control}
\label{sec:approach}

In this section we describe the reward shaping strategy proposed to enable learning (\sref{sec:rewards}); the continuous and discrete actions baseline \gls{rl} methods considered (\sref{subsec:dqn} and \sref{subsec:acmethods}); and the sim-to-real approach taken to reduce the reality gap (\sref{subsec:sim2real}).

\subsection{Reward Shaping for Soccer Agents}
\label{sec:rewards}
As our results will show, using only the natural rewards of the task $R_g$ (goals scored: +1 if the team scores a goal or -1 if the other team scores) is not sufficient for training the agents, due to the sparsity of goal events. Therefore, we added three per step reward components to the natural reward that guide the agent to learn the objective of the task. The first component rewards the motion towards the ball, $R_m$, at time $t$ and time step $dt$, given by:

\begin{equation}
    R_m = \frac{d(a, b)_t - d(a, b)_{t-dt}}{dt},
\end{equation} 
where $d$ is the euclidean distance, and $a$ and $b$ are the positions of the agent and ball, respectively. It rewards the agent positively if its distance from the ball decreases, and negatively otherwise. 

The second is the ball position gradient component, $R_p$: 
\begin{equation}
        R_p = \frac{bp_t - bp_{t-dt}}{dt},
\end{equation}
\begin{equation}
        bp = \frac{\frac{d(g_o, b) - d(g_a, b)}{170} - 1}{2},
\end{equation}
where $g_o$ and $g_a$ are the positions of the center of the own goalpost and adversary goalpost, respectively. This component is defined as the discrete derivative of the difference between the ball's distances relative to each goalpost. It rewards the agent for interacting with the ball moving it towards the opponent's goal. 

Third, the energy component ($R_e$) is used to penalize energy usage and is given by $R_e = -(|v_l| + |v_r|),$
where $v_l$ and $v_r$ are the linear velocities of the left and right wheels, respectively.

The goal ($R_g$), motion ($R_m$), ball position gradient ($R_p$) and energy ($R_e$) rewards are compose by a weighted sum to form the reward given for the the agents at every step: $R = w_g R_g + w_m R_m + w_p R_p + w_e R_e$,
where the weights are parameters set to the following values by trial and error: $w_g = 1.0$, $w_m = 0.02$, $w_p = 0.08$, and $w_e = 1^{-5}$ for the continuous actions method and $w_e = 0$ for the discrete one.

In \citep{ng1999policy}, the authors prove that using potential-based functions for reward shaping does not modify the optimal policy obtained by the agents. Therefore, all the components proposed above follow this format.

\subsection{Discrete Actions Baseline}
\label{subsec:dqn}

We chose \gls{dqn}\citep{volodymyr2013playing} to train our discrete actions agent due to its wide application in the \gls{rl} as a baseline method. The selected action needs to be converted to the continuous domain for our application. In the proposed approach, the DQN agent controls its desired position by moving a virtual target in the field. The way the agent will reach the desired target is controlled by a fixed low-level control that is responsible for moving the agent to the desired position. The learned policy can handle only with the expected behavior in the field. The DQN agent then selects one of five actions that change the target position in a polar coordinate system with respect to agent position: target remains at same position ($a_1$); target is rotated by $\pm 15$ degrees clockwise ($a_2$) or counterclockwise ($a_3$); target distance is increased ($a_4$) or decreased ($a_5$) by 12 centimeters.
This approach is similar to what was proposed in \citep{duan2007application}, serving as a way to compare our results with previous work.

\subsection{Continuous Actions Baselines}
\label{subsec:acmethods}

To evaluate our continuous actions agent we chose two state-of-the art \gls{ac} methods: \gls{ddpg}\citep{lillicrap2015continuous} and \gls{sac}\citep{sac}.
The agent in the continuous domain is controlled through its desired linear and angular velocities. Then, the direct kinematics of the robot is used to derive the wheels velocities. This control approach was chosen to minimize the consequences of mistakes. Since the wheels velocities are highly dependent on each other, small errors could lead to undesired motions.

\subsection{Sim-to-Real: Low-Level Control with a Feed Forward Network}
\label{subsec:sim2real}

The proposed transfer approach is based on Domain Adaptation. It creates a low-level control abstraction layer that provides environment-robot independence for the high-level control policy. This allows low-level control reuse for different policies, reducing the training time by avoiding the need for sample hungry domain randomization techniques. The training process is illustrated in \fref{fig:sim2realtrain}.

Consider $a_{d}(t) = \{v, \omega\}$, a pair of linear and angular speeds, as the action desired by the policy at time $t$ and $a_{o}(t)$ the resulting action observed in the real world. Consider also that exists another action in the target environment $a_{r}(t)$, wheel speeds, that would approximate the expected result of the desired action ($a_{r}(t) \sim a_{d}(t)$). Thus, a function $F(a_{d}(t)) = a_{r}(t)$ must be learned.
We learn obtain $F$ by learning the the inverse dynamics model of the agent-environment, also taking into account the action executed in the previous step, $a_{r}(t-1)$, using $F(a_{d}(t), a_{r}(t-1)) = a_{r}(t)$. This approach is similar to \citep{christiano2016transfer}, but does not require to run the simulation at every step of the real environment.

In this work, the function $F$ is learned using a feed-forward neural network fed with data (trajectories) collected from the real world. The low-level control produces the wheel speeds $V_L$ and $V_R$ as $a_{r}(t)$. The collected trajectories are composed of the measured history of robot angular and linear speeds, followed by the wheels speeds that produced these velocities. Therefore, the neural network learns which wheel speeds produce the desired linear and angular speeds.

\section{Experimental Results}
\label{sec:results}

In this section we discuss the obtained results in simulation (\sref{subsec:results_sim}), an evaluation of the sim-to-real approach (\sref{subsec:results_real}), and a comparison of the obtained policies with policies designed by humans (\sref{subsec:results_game}).

\subsection{Results with DRL Algorithms in Simulation}
\label{subsec:results_sim}

The agents trained in the simulation are evaluated based on their goal score, which translates to the aptitude of an agent to score goals in the opponent's goalpost and avoid goals in its own goalpost. The agent's goal score is defined by the agent's accumulated goals reward in the window of the next hundred steps. Each method shown in the graphs was trained ten times with a different random initialization.

\begin{figure*}[ht!]
  \centering
  \subfigure[]{\includegraphics[width=0.49\linewidth,scale=1]{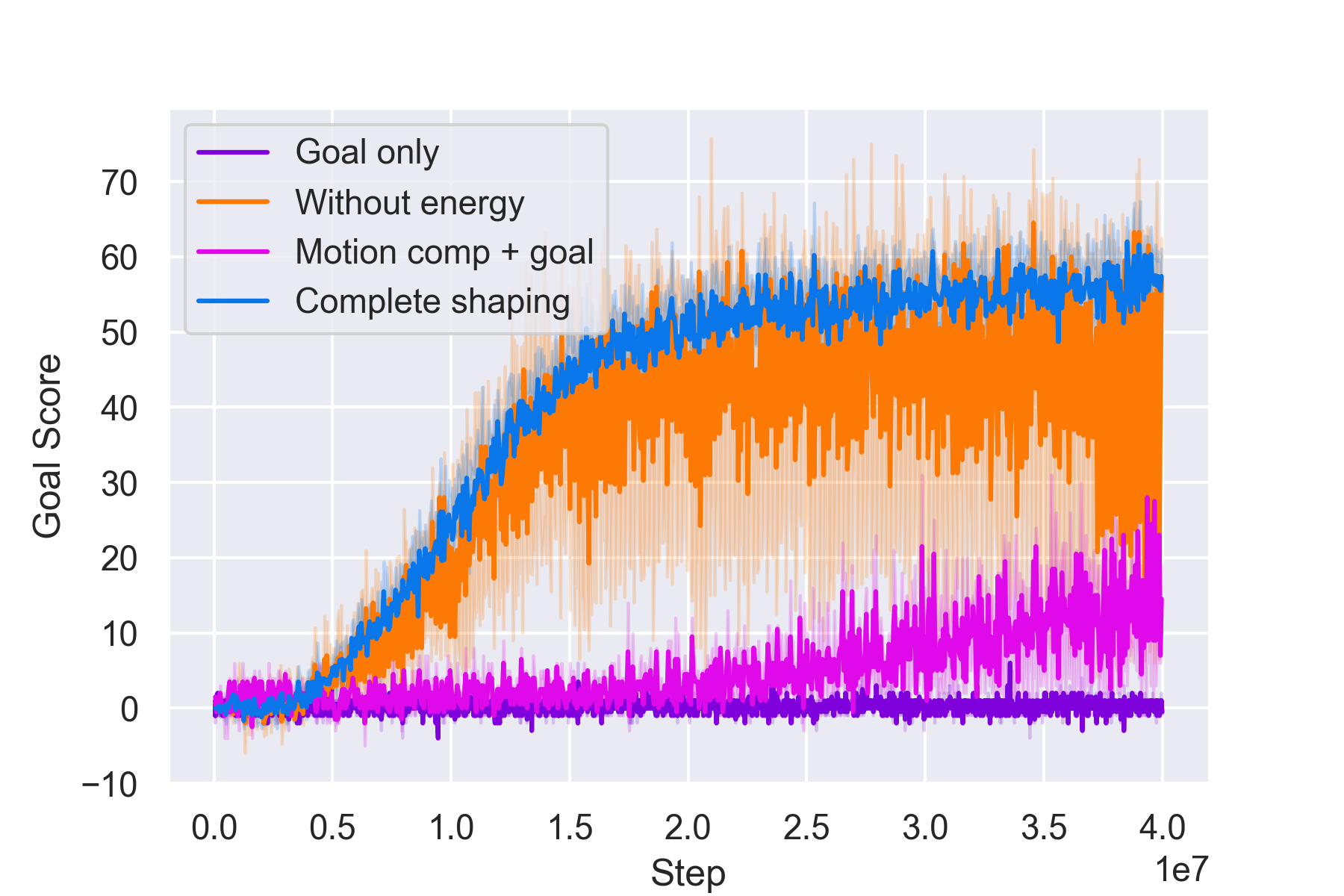}
  \label{fig:rew_shaping}}
      \subfigure[]{\includegraphics[width=0.49\linewidth,scale=1]{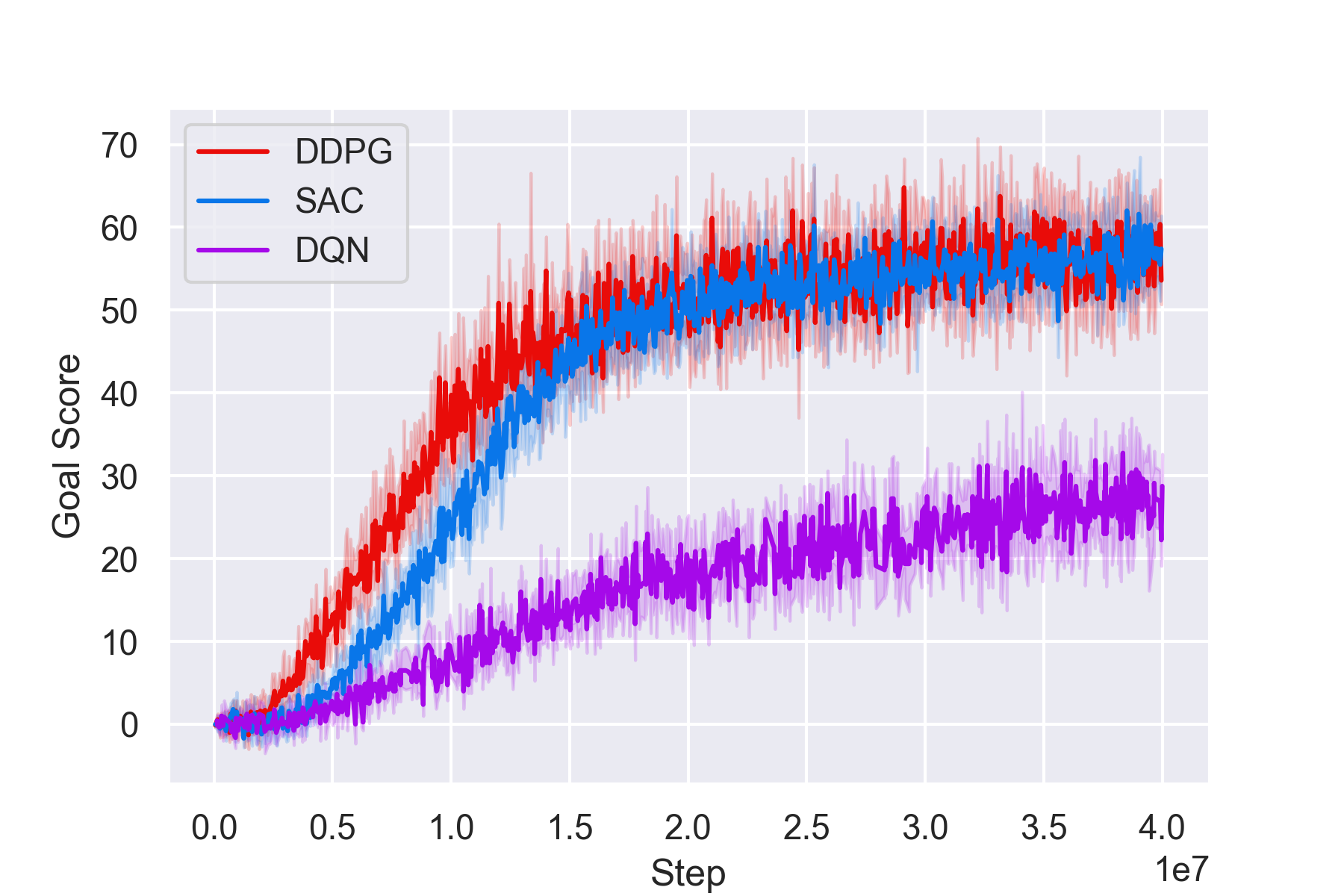}
  \label{fig:allmodels}}
    \caption{(a) Impact of reward shaping for SAC. Though not shown, the results with DDPG are similar. (b) Comparison between DQN, DDPG, and SAC baselines.}
  \label{fig:plots}
\end{figure*}

The importance of reward shaping is illustrated in \fref{fig:rew_shaping}. The \gls{sac} agent with only sparse goals rewards does not learn any desired behavior. Adding the motion reward enables some learning, though slow. When the ball position gradient was added, the agent learns faster to push the ball towards the opponent's goal, rapidly increasing the goal score, but with unstable results. When the energy component was added, the learning becomes more stable.

In \fref{fig:allmodels}, we can observe that, with reward shaping, all baselines were capable of learning intelligent behaviors. \gls{sac} and \gls{ddpg} baselines outperform \gls{dqn} by approximately fifty percent in goal score efficiency. The continuous control method better suited since it directly controls linear and angular velocities, being able to change them faster. \gls{dqn}, on the other hand, must control the velocities through the integration of the agent's target position and is bounded to follow the path dictated by its rigid low-level control method.

\subsection{Evaluation of the Sim-to-Real Approach}
\label{subsec:results_real}

We evaluate the sim-to-real transfer by executing the policy in simulation and in the real environment with and without the low-level control with the neural network and compare the average steps to score a goal achieved by each method (the lower, the better). Each method runs for ten episodes, and each episode ends with a goal or in 5 minutes of playing. We used the policy obtained with \gls{dqn} to evaluate the transfer learning.

In the simulation scenario, the agent takes $547.2 \pm 233.6$ steps to score a goal. In the real environment, without sim-to-real, the agent takes $901.1 \pm 422.9$ steps, while with sim-to-real, it takes only $456.8 \pm 147.2$ steps. The results indicate a considerable improvement in performance in the real world by using the proposed method.

Sim-to-real allowed us to transfer the policies learned in simulation to the real-world environment without retraining them. By applying the proposed sim-to-real approach, the performance obtained in the real world is statistically equivalent to the one observed in simulation (p-value = 0.3101). It is important to highlight that the real robot has a concave frontal shape that improves its ability to carry the ball compared to the simulated robot. This factor should also be considered in this result.



\subsection{Comparison with Policies Designed by Humans}
\label{subsec:results_game}

To evaluate the policy learned with \gls{ddpg} in a more realistic competition scenario, we invited the VSSS team, third place on the LARC 2018, for a 1-vs-1 game. The team used the latest version of their striker policy. It employs univector fields \citep{kim2004humanoid} for path planning and low-level motion control with a PID, capable of achieving high speeds while maintaining precise control. It also has a high-level decision module that identifies the current situation and switches behaviors among a predefined repertoire set, including spinning, approach ball, and carry the ball to the goal.

Two games of ten minutes each were performed, and the goals were registered both manually and automatically. Invalid goals were discarded, and a few interventions were made when the robots stuck to avoid damaging the motors.

The final scores of the matches were 19 for the \gls{ddpg} agent and 13 for the opponent team in the first game, and 22 for the \gls{ddpg} approach and 17 for the opponent team in the second. These results confirm the superiority of the proposed approach in the single-agent scenarios.

As can be observed in the supplementary video, the movements of the opponent striker are faster and more precise, while the \gls{ddpg} agent displays a much broader repertoire of behaviors.

The final evaluation of the proposed approach was done in \gls{larc} 2019. The single-agent policy obtained with \gls{ddpg} was used to control the three agents of a complete team in the \gls{vsss} competition. \tref{tab:larc2019} presents the scores of each match in chronological order. With these results, our team was ranked fourth place in the competition, confirming that the policy obtained was competitive even against the best teams in the league.

\begin{table}[H]
\centering
\caption{Scores of each match in the real-world competition between the proposed model and an opponent. In bold, are the games that finished before the match time due to $10$ goals difference rule.}
\label{tab:larc2019}
\resizebox{0.4\textwidth}{!}{%
\begin{tabular}{ccccc}
\toprule
\textbf{VSSS-RL} & \textbf{$10$} & \textbf{$\times$} & \textbf{$00$} & \textbf{Team $1$  (5th Place)} \\
\textbf{VSSS-RL} & \textbf{$10$} & \textbf{$\times$} & \textbf{$00$} &  \textbf{Team $2$ (13th Place)} \\
\textbf{VSSS-RL} & \textbf{$11$} & \textbf{$\times$} & \textbf{$01$} &  \textbf{Team $1$ (5th  Place)} \\
\textbf{VSSS-RL} & \textbf{$11$} & \textbf{$\times$} & \textbf{$01$} &  \textbf{Team $2$ (13th Place)} \\
\textbf{VSSS-RL} & \textbf{$10$} & \textbf{$\times$} & \textbf{$00$} &  \textbf{Team $3$ (13th Place)} \\
\textbf{VSSS-RL} & \textbf{$11$} & \textbf{$\times$} & \textbf{$01$} &  \textbf{Team $3$  (13th Place)} \\
VSSS-RL          & $08$          & $\times$          & $00$          & Team $4$           (7th Place)          \\
VSSS-RL          & $05$          & $\times$          & $07$          & Team $5$           (2nd Place)          \\
VSSS-RL          & $09$          & $\times$          & $05$          & Team $6$           (3rd Place)          \\
\textbf{VSSS-RL} & \textbf{$14$} & \textbf{$\times$} & \textbf{$04$} & \textbf{Team $1$  (5th Place}) \\
VSSS-RL          & $06$          & $\times$          & $07$          & Team $6$           (3rd Place)         \\
\bottomrule
\end{tabular}%
}
\end{table}

\section{Conclusions and Future Work}
\label{sec:conclusion}

In this work, we proposed a framework for training robots for the \gls{vsss} league. It can be used for research in single-agent \gls{rl}, \gls{marl} and sim-to-real methods with support for self-play. The environment is fast and stable, and the real robots for the \gls{vsss} league are cheap and easy to build and maintain. Moreover, the policies obtained can be evaluated yearly in the real world at the RoboCup competitions against the best teams. We also point out that reproducibility was easily achieved in the simulated environment, regardless of initialization, and most hyper-parameters do not affect the results significantly, making it a good candidate for benchmarking \gls{rl} and \gls{marl} methods. We believe that these features make VSSS-RL an excellent tool for evaluating methods designed for competition and collaboration in dynamic environments.

The reward shaping function proposed allowed us to successfully train in simulation three off-policy methods for controlling the desired linear and angular speeds of the \gls{vsss} robots. \gls{sac} and \gls{ddpg} displayed better results than \gls{dqn} due to their ability to specify precisely the robot speeds through continuous outputs values. Both methods achieved similar results on average, though \gls{sac} was more stable throughout the runs.

The behaviors displayed by the policies learned by the \gls{rl} are rich and complex, challenging to specify by hand and  to identify the correct situations when they should be applied. For instance, we can highlight the behaviors of pushing and blocking the opponent, using the sides of the robot to guide the ball, combine linear and angular speeds to kick the ball in the right direction, and bounce the ball on the walls. These behaviors can be observed in the supplementary video provided both in simulation and the real world. However, we observed that the trajectories planned by the \gls{rl} baselines are frequently short-sighted and reactive, for instance, not taking into account probable collisions at the beginning of the movements reacting fast when they occur.


The results in the 1-vs-1 scenario and the 4th place obtained in the last edition of \gls{larc}, with bold victories obtained against traditional teams, can be seen as an emblematic example of successful application of \gls{rl} in the real world. To the best of our knowledge, this is the first time that a policy trained by \gls{rl} to control the complete behavior of a soccer robot has won against policies designed by humans in a competition. 

The facts that our team was not the first place and that we were not able to train multi-agent policies, indicate that there is still space for future research. We believe that better results could be achieved by incorporating roll-outs collected in the real environment using off-policy methods for fine-tuning. We also plan to evaluate on-policy \gls{rl} methods such as Proximal Policy Optimization \citep{schulman2017proximal} and Trust Region Policy Optimization \citep{schulman2015trust} and implement \gls{marl} methods, such as Multi-Agent \gls{ddpg} \citep{lowe2017multi}, to train a complete team for competing in the next \gls{larc}.

\section*{ACKNOWLEDGMENTS}

The authors would like to thank RoboCIn - UFPE Team and Mila - Quebec Artificial Intelligence Institute for the collaboration and resources provided; Conselho Nacional de Desenvolvimento Cientifico e Tecnológico (CNPq), and Coordenação de Aperfeiçoamento de Pessoal de Nível Superior (CAPES) for financial support. Moreover, the authors also gratefully acknowledge the support of NVIDIA Corporation with the donation of the Titan V GPU used for this research.

\bibliography{default_content/bibliography}

\end{document}